\documentclass[conference]{./IEEEtran}
\IEEEoverridecommandlockouts
\usepackage[utf8]{inputenc}
\usepackage{cite}
\usepackage{amsmath,amssymb,amsfonts}
\usepackage{algorithmic}
\usepackage{graphicx}
\usepackage{textcomp}
\usepackage{xcolor}
\def\BibTeX{{\rm B\kern-.05em{\sc i\kern-.025em b}\kern-.08em
    T\kern-.1667em\lower.7ex\hbox{E}\kern-.125emX}}

\usepackage{bm}
\usepackage{bbm}
\usepackage{dsfont}
\usepackage{siunitx}
\usepackage{multirow}
\usepackage{booktabs}
\usepackage{tabularx}
\newcolumntype{Y}{>{\centering\arraybackslash}X} 	

\usepackage{caption}
\DeclareMathOperator*{\argmax}{arg\,max}
\usepackage{tikz}
\usepackage{pgfplots}					
\usepackage{pgfplotstable}
\usetikzlibrary{decorations}

\pgfplotsset{width=7cm,compat=newest, every tick label/.append style={font=\tiny}}	
\usepackage{filecontents}
\usepackage{pdfpages}
\usepackage{pgf}
\usepackage{nicefrac}
\usepackage{cite} 	

\newcommand{\bfvscore}{$\mathbf{V_1}$}
\newcommand{\bffscore}{$\mathbf{F_1}$}
\newcommand{\dsA}{$\mathbb{S}_1$}
\newcommand{\dsB}{$\mathbb{S}_2$}
\newcommand{\yaw}{\phi} 
\newcommand{\radvel}{v_r}
\newcommand*{\abs}[1]{\left\lvert#1\right\rvert}		
\begin{document}

\title{Off-the-shelf sensor vs. experimental radar --\\How much resolution is necessary in\\automotive radar classification? \thanks{This research received funding from the European Union under the H2020 ECSEL program as part of the DENSE project, contract number 692449.}}

\author{\IEEEauthorblockN{Nicolas Scheiner\IEEEauthorrefmark{1},
Ole Schumann\IEEEauthorrefmark{1},
Florian Kraus\IEEEauthorrefmark{1},
Nils Appenrodt\IEEEauthorrefmark{1},
Jürgen Dickmann\IEEEauthorrefmark{1}, and 
Bernhard Sick\IEEEauthorrefmark{2}}
\IEEEauthorblockA{\IEEEauthorrefmark{1}\textit{Environment Perception},
\textit{Mercedes-Benz AG},
Stuttgart, Germany, 
Email: nicolas.scheiner@daimler.com}
\IEEEauthorblockA{\IEEEauthorrefmark{2}\textit{Intelligent Embedded Systems},
\textit{University of Kassel},
Kassel, Germany,
Email: bsick@uni-kassel.de}
}

\maketitle
\thispagestyle{plain}
\pagestyle{plain}

\begin{abstract}
Radar-based road user detection is an important topic in the context of autonomous driving applications.
The resolution of conventional automotive radar sensors results in a sparse data representation which is tough to refine during subsequent signal processing.
On the other hand, a new sensor generation is waiting in the wings for its application in this challenging field.
In this article, two sensors of different radar generations are evaluated against each other.
The evaluation criterion is the performance on moving road user object detection and classification tasks.
To this end, two data sets originating from an off-the-shelf radar and a high resolution next generation radar are compared.
Special attention is given on how the two data sets are assembled in order to make them comparable.
The utilized object detector consists of a clustering algorithm, a feature extraction module, and a recurrent neural network ensemble for classification.
For the assessment, all components are evaluated both individually and, for the first time, as a whole.
This allows for indicating where overall performance improvements have their origin in the pipeline.
Furthermore, the generalization capabilities of both data sets are evaluated and important comparison metrics for automotive radar object detection are discussed.
Results show clear benefits of the next generation radar.
Interestingly, those benefits do not actually occur due to better performance at the classification stage, but rather because of the vast improvements at the clustering stage.
\end{abstract}



\section{Introduction}
\begin{figure}[t!]
	\vspace{-8pt}
	\centering
	\includegraphics[width=0.99\linewidth]{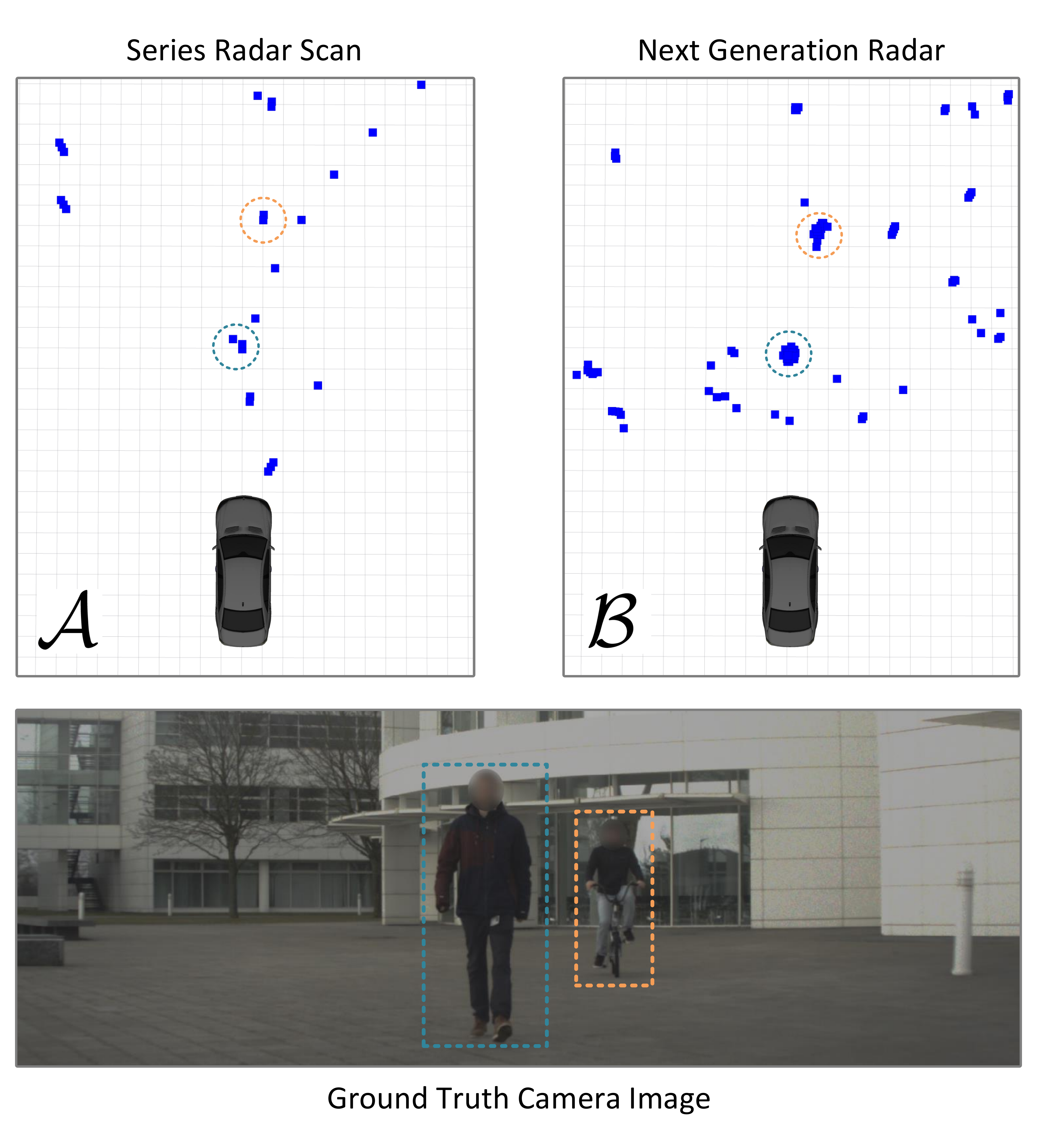}
	\vspace{-22pt}
	\caption{Two radar sensor recordings and a camera image of the same scene. On the left, the radar detections of the off-the-shelf radar are visualized. On the right, the same scene is captured by a next generation radar. Through visual augmentation, the superiority of the next generation radar is clearly visible as much more detection points fall on the two objects of interest. The higher amount of clutter points is far less stable than the real object detections and can be easily distinguished from the real objects in moving scenarios. In this article, the general validity of this superiority for machine learning algorithms is investigated.}
	\label{fig:graphical_abstract}
	\vspace{-15pt}
\end{figure}
Driven by market demands for advanced driver assistance systems and autonomous vehicles, automotive radar sensors are continuously evolving.
One important step in their development was the utilization of printed-circuit-board antenna arrays.
By abandoning waveguides and mechanically steered antennas, the sensors can be manufactured with more antenna elements while being cheaper and more compact than their predecessors.
Nowadays, there is a trend towards using more and more antenna elements in a sensor to increase the radar's angular ambiguity range.
Moreover, additional elements can be used to measure not only the incident angle of a reflecting signal in azimuth, but also in elevation, or even to measure polarimetric information \cite{Weishaupt2019, Tilly2019}.
Beside the antennas, other internal hardware components are also continuously evolving.
Hence, it is now possible to utilize, for example, higher sampling rates or steeper frequency ramps, leading to an improved sensor resolution.

It was shown, e.g. in \cite{Brisken2019} and \cite{Schubert2015a}, that these improved sensor specifications can serve well for the discrimination between two target objects and even allow for discerning body parts of pedestrians \cite{Steinhauser2019}.
Furthermore, current research showed that automotive object detection tasks can be eased using these new types of radar sensors \cite{Meyer2019}.

Probably due to the lack of publicly available data sets, a systematic evaluation with state-of-the-art classifiers on real world automotive radar scenes is currently not available.
The necessity for such an evaluation is apparent for a simple reason.
Every new generation of radar sensors produces a lot of work and costs, starting from the development, the integration in a car, and finally the collection of new data sets along with extensive testing.
A major concern in this process is that even though the data might seem superior to the data engineers, e.g., as visible in Fig.~\ref{fig:graphical_abstract}, an object detector may actually not benefit from it.
Modern classifiers for automotive radar data are often based on machine learning methods, e.g. \cite{Scheiner2019IV, Schumann2018, Schumann2019, Danzer2019, Lombacher2017, Palffy2020}.
For these methods it is more complicated to predict in what way the algorithms actually benefit from higher resolved data.

In this article, the preconception, ``\emph{high resolution sensors are always better,}'' is investigated.
Therefore, two data sets from two different cars are put together in a manner that allows for a fair comparison later on.
Moreover, a modularized object detector consisting of a clustering algorithm, a feature extractor, and a recurrent neural network ensemble is described and optimized separately on both sensor data sets.
The object detection results are evaluated based on the respective sensor.
Finally, an estimation on the quantitative benefits of using a next generation radar is presented for all components of the object detector.
In addition to a sensor comparison, the combined assessment of recently published object detection modules allows for the first in-depth investigation of the interaction of those modules.

In summary, the following contributions are made:
\begin{itemize}
	\item Two sensors from different generations are evaluated using real world data recordings.
	\item Every detection module is compared individually to highlight the exact sensor advantages.
	\item The utilized object detector framework is presented as a whole for the first time.
	\item Important assessment factors are identified to ease the evaluation for future radar generations.
\end{itemize}

\section{Data Sets} \label{sec:datasets}
For the comparison of the two radar sensors, two different proprietary data sets are used.
Both contain recordings from different real world scenes and were collected with two individual test vehicles.
Please note, that there are no open radar data sets which can be used for the purposes of this article.
The only currently available data sets comprise nuScenes \cite{nuscenes2019}, Astyx \cite{astyxDataset}, and Oxford Radar RobotCar \cite{RadarRobotCarDatasetICRA2020}.
However, the nuScenes radar has far worse data density even than the off-the-shelf sensor compared here.
The Astyx data set has a much higher data density, but it is too small overall.
Lastly, the Oxford Radar RobotCar data set comprises automotive scenarios but does not use an automotive radar sensor which fails the purpose of this article.

In this section, an overview is given of the vehicle setups, the sensor specifications, and how the recorded scenes of both data sets compare.
The first data set, recorded by vehicle $\mathcal{A}$ using four off-the-shelf radar sensors is referred to as $\mathbb{S}_1$ and the data set recorded by vehicle $\mathcal{B}$ with two experimental next generation radars as $\mathbb{S}_2$.
The sensor specifications for both sensors are listed in Tab.~\ref{tab:sensor_specs}.
The upper half of the table represents the frequency range $f$ of the emitted signal and the operational bands for range (distance) $r$, azimuth angle $\yaw$, and radial (Doppler) velocity $\radvel$ respectively.
In the second part the resolutions  $\Delta_r, \Delta_\yaw, \Delta_{\radvel}$ and $\Delta_t$ for $r$, $\yaw$, $\radvel$, and time $t$  are noted.
All values are given separately for both sensors, the conventional sensor in vehicle $\mathcal{A}$ and the next generation sensor in vehicle $\mathcal{B}$.
Both sensors operate at the same frequency range.
Sensors in vehicle $\mathcal{A}$ have a wider Doppler velocity ambiguity range and a faster sensor update rate, i.e., a higher time resolution.
The latter is actually not a sensor property but a practical choice during recording with vehicle $\mathcal{B}$ in order to decrease the data load.
Theoretically, the maximum update rate for the next generation radar lies at \SI{20}{\hertz}.
In all other categories, the experimental sensors in vehicle $\mathcal{B}$ are superior to the off-the-shelf sensors in $\mathcal{A}$.
For the recorded scenarios, the most important aspects are the two to five times smaller resolution values for range, angle, and Doppler.
The detection filtering thresholds have been set to well-fitting thresholds for both sensors types, i.e., to find a good compromise between false positive and false negative detection. Hence, no sensor is favored over the other. 
\begin{table}[tb]
	\renewcommand{\arraystretch}{1.3}
	\caption{Radar sensor specification for both compared sensors.}
	\label{tab:sensor_specs}
	\centering
	\begin{tabular}{ccccc}
		Sensor & $f / \SI{}{\giga\hertz}$ & $r / \SI{}{\meter}$ & $\yaw / \deg$ & $\radvel /\SI{}{\meter\per\second}$\\
		\midrule
		$\mathbf{\mathcal{A}}$ & $76-77$ & $0.25-100$ & $\pm60$ & $-111 - +222$ \\
		$\mathbf{\mathcal{B}}$ & $76-77$ & $0.15-153$ & $\pm70$ & $-44.3 - +44.3$ \\
		\bottomrule\\
		Sensor & $\Delta_t / \SI{}{\milli\second}$ & $\Delta_r / \SI{}{\meter}$ & $\Delta_\yaw /\deg$ & $\Delta_{\radvel} /\SI{}{\meter\per\second}$\\
		\midrule
		$\mathbf{\mathcal{A}}$ & $60$ & $0.42$ & $3.2 - 12.3$ & $0.43$\\
		$\mathbf{\mathcal{B}}$ & $100$ & $0.15$ & $1.8$ & $0.087$\\
		\bottomrule
	\end{tabular}
\end{table}

\begin{figure}
	\centering
	\includegraphics[width=0.55 \linewidth]{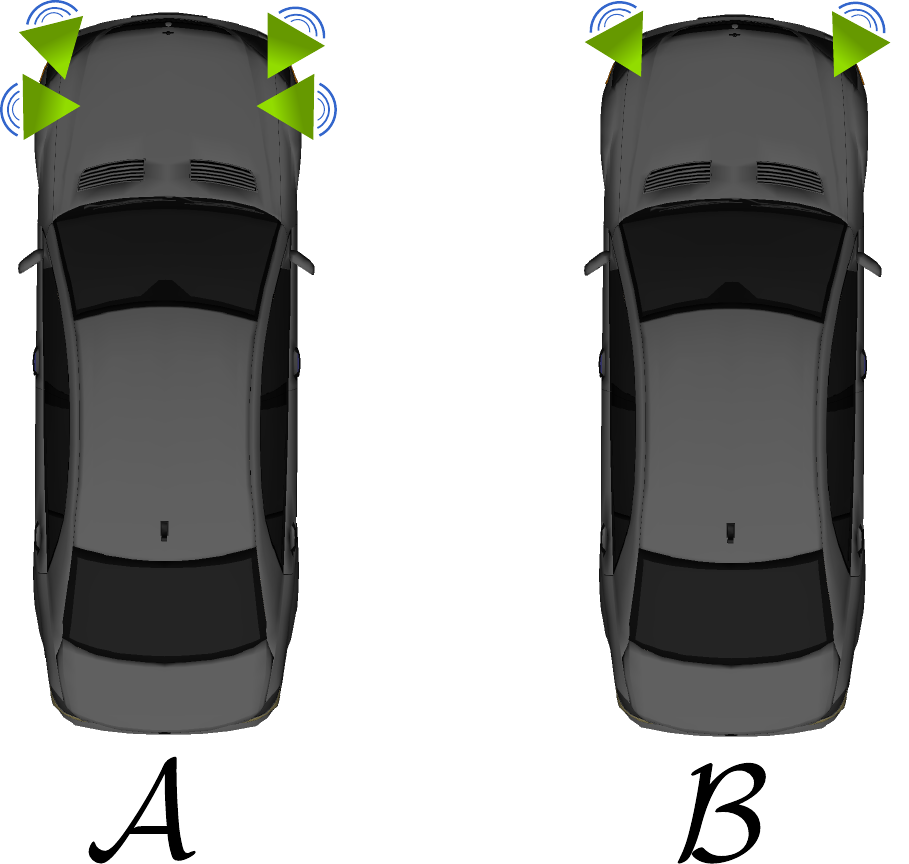}
	\caption{Schematic sensor distribution of both test vehicles. Test car $\mathcal{A}$ was used to collect the data for $\mathbb{S}_1$. $\mathbb{S}_2$ was recorded with car $\mathcal{B}$. Both vehicles contain only a single radar sensor type.}
	\label{fig:vehicle_setup}
\end{figure}

The sensor setups for both test vehicles are displayed in Fig.~\ref{fig:vehicle_setup}. 
For $\mathbb{S}_1$, four sensors are distributed over the front bumper of the car.
For $\mathbb{S}_2$, however, only two sensors are mounted in the front of the vehicle, as their wider field of view suffices for a coverage of the entire vehicle front.
Due to data sparsity, overlapping regions in the sensors' fields of view can be superposed by simple data accumulation.
In order to keep the sensors from interfering with each other, the sensor cycles are interleaved.

Both, $\mathbb{S}_1$ and $\mathbb{S}_2$, are subsets of larger proprietary data sets.
In order to make both data sets comparable, special attention has to be payed to the selection of appropriate recordings.
Especially for vehicle $\mathcal{B}$, the original data base has a strong imbalance between classes and currently contains only very few examples of, e.g., cars, trucks, and motorcycles.
To make both data sets comparable, solely recordings with pedestrians, bicycles, and background detections are used for the evaluation in this article.
The background class consists of static points, measurement artifacts, and other clutter.
Furthermore, vehicle $\mathcal{B}$ remains stationary during a large percentage of the recordings, i.e., no ego-motion compensation is necessary.
This is a potentially advantageous factor for later classification, and, therefore, for both data sets only data from scenarios without ego-motion are chosen.

This results in a total of 53 selected sequences for data set $\mathbb{S}_1$ with a total length of about \SI{76}{\minute}. 
All data was manually labeled by human experts.
The second data set $\mathbb{S}_2$ contains 25 sequences with a total of 105 repetitions, i.e., many sequences have been recorded multiple times at the same locations, but with different road user trajectories.
The recording time of $\mathbb{S}_2$ adds to about \SI{67}{\minute}.
$\mathbb{S}_2$ consists of 5 purely manually labeled recordings and 100 further ones, where a hand-held global navigation satellite system reference was used for automatic labeling according to \cite{Scheiner2019IRS}.
All automatically labeled data were manually checked and corrected if necessary.
The distributions of object samples and detection points within both data sets are given in Tab.~\ref{tab:data}.
``Objects'' refers to the amount of \SI{150}{\milli\second} time windows during which the actual object instances are present in the data.
For the background class, object samples are created by removing all ground truth objects from the data and afterwards clustering the remainder of the detection points with the same clustering algorithm as discussed in Sec.~\ref{sec:meth_cluster}.
The reported numbers are obtained after application of the data filter discussed also in Sec.~\ref{sec:meth_cluster}, which basically reduces the amount of background detections to roughly one tenth.
One important observation from Tab.~\ref{tab:data} is the strong data imbalance between the classes in both data sets.
Set $\mathbb{S}_1$ is a lot larger with respect to object samples.
However, the amount of detections on the actual road users is similar, i.e., the amount of detections per road user is much larger for the next generation sensor in $\mathbb{S}_2$ due to the higher sensor resolution.
Despite the remaining differences between both data sets, their sizable quantities together with the described sequence selection strategy allows for a suitable comparison.
\begin{table}[tb]
	\renewcommand{\arraystretch}{1.3}
	\caption{Data set distribution comparison after filtering. For each data set and class, the amounts of object samples and detection points are listed.}
	\label{tab:data}
	\centering
	\begin{tabular}{crccc}
		\toprule
		Data Set & & Pedestrian & Bicycle & Static/Garbage \\
		\midrule
		\multirow{2}{*}{$\mathbb{S}_1$} & Objects & 22424 & 5810 & 66020 \\
		& Detections & $2.96\cdot 10^5$ & $1.41\cdot 10^5$ & $8.09\cdot 10^5$ \\
		\midrule
		\multirow{2}{*}{$\mathbb{S}_2$} & Objects & 2751 & 1809 & 13248 \\
		& Detections & $2.58\cdot 10^5$ & $2.27\cdot 10^5$ & $1.14\cdot 10^{10}$\\
		\bottomrule
	\end{tabular}
\end{table}

\section{Comparison Methodology}
\begin{figure*}[tb]
	\centering
	\includegraphics[width=0.9\linewidth]{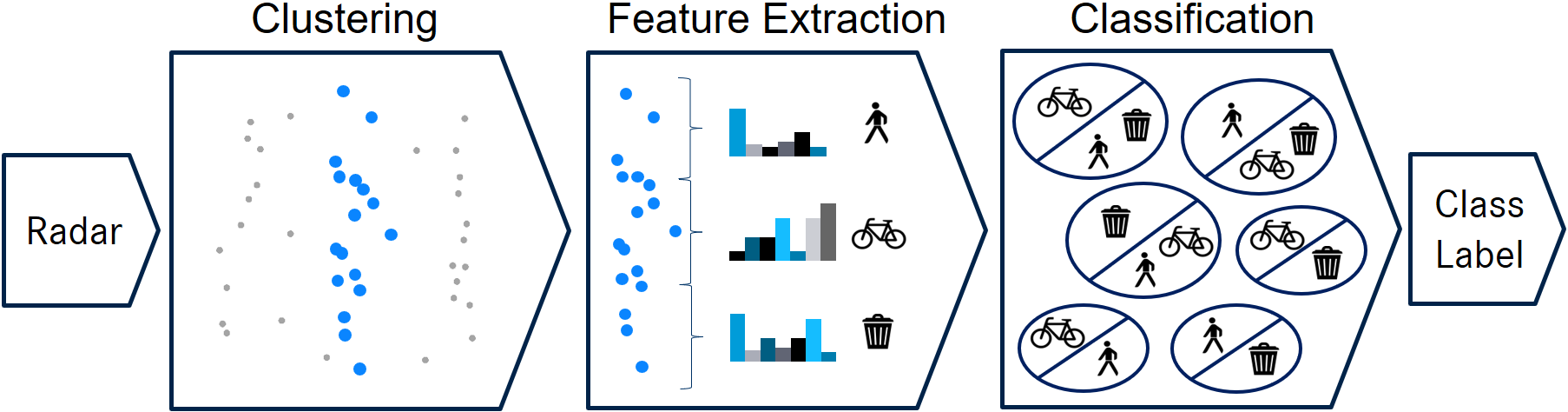}
	\caption{Modularized object detection framework: The data are first structured using a clustering algorithm. Every cluster formation is subject to feature extraction. For each cluster, six feature sets are extracted -- three of those are visualized. Each feature set is optimized for one of the six classifiers in the classification stage. Recurrent neural networks with a single layer of long short-term memory (LSTM) cells are used as classifiers. The set of classifiers consists of three one-vs-all classifiers and three one-vs-one classifiers which make a combined final class decision for each cluster sample.}
	\label{fig:pipeline}
\end{figure*}

A basic overview of the utilized detection framework is given in Fig.~\ref{fig:pipeline}.
The pipeline consists of three stages which expect as input radar detection points which are resolved in range $r$, azimuth angle $\phi$, Doppler velocity $v_r$, and time $t$.
Furthermore, for each detection an amplitude is measured which is an estimation of the radar cross section of the part of the object the detection belongs to.
The main components of the framework serve to:
\begin{enumerate}
	\item[A.] Clustering: merge radar detections to object instances
	\item[B.] Feature extraction: enrich the feature space of the data clusters by collecting cluster meta information
	\item[C.] Classification: make a class decision for each cluster
\end{enumerate} 
The remainder of this section gives details about the three parts of the algorithm.

\subsection{Data Clustering} \label{sec:meth_cluster}
The clustering method uses a DBSCAN algorithm \cite{Ester1996}, following the findings from \cite{Scheiner2019ITSC}.
First, the data points are transformed to Cartesian coordinates $x$ and $y$, and pre-filtered in order to ease the actual clustering process.
For filtering, the clustering algorithm is altered to consider only points above a certain velocity threshold which depends on the number of detection points in a close distance $d_{xy}$ according to:
\begin{equation}\label{eq:filter_metric}
\begin{split}
\abs{v_r} < \mathbf{\eta}_{\mathbf{v}_{\mathbf{r},i}} \:\wedge\: N(d_{xy}) < \mathbf{N}_i \\
\quad \text{with } \mathbf{N} \in \mathbb{N}^5, \: \mathbf{\eta}_{\mathbf{v}_{\mathbf{r}}} \in \mathbb{R}^5.
\end{split}
\end{equation}
The neighbor threshold $\mathbf{N}$, the velocity threshold $\mathbf{\eta}_{\mathbf{v}_{\mathbf{r}}}$, and the spatial search radius $d_{xy}$ are parameters which have to be optimized.

The other differences to conventional DBSCAN are an adaptive number of minimum points $N_{\text{min}}(r)$ required to form a cluster core point, where $r$ is the range, i.e., distance between detection and sensor.
Therefore, the sensor-specific range information of each detection has to be kept in the data to avoid extra calculations.
This adjustment is based on the fact that remote objects have a smaller maximum number of possible detections due to range-independent angular resolution as discussed in \cite{Kellner2012}.
Since the physical extents of road users do not change, the minimum point property of DBSCAN is set to:
\begin{equation} \label{eq:min_pts}
\resizebox{.89\linewidth}{!}{%
	$
	N_{\text{min}}(r) = N_{\text{min},\SI{50}{\meter}} \cdot \left(1 + \alpha_r \cdot \left(\frac{\SI{50}{\meter}}{\text{clip}(r,\SI{25}{\meter},\SI{125}{\meter})}-1\right)\right).
	$}
\end{equation}
Both $N_{\text{min},\SI{50}{\meter}}$ and $\alpha_r$ are tuning parameters which represent a minimum point baseline at $\SI{50}{\meter}$ and the slope of the reciprocal relation.
Furthermore, only detections that exceed a certain radial velocity threshold $\abs{v_r} > v_{r,\text{min}}$ can become \emph{core points} in accordance with \cite{Schumann2018b}.

Lastly, the distance region (also known as $\epsilon$ region) of the DBSCAN algorithm is customized to cluster points that have low spatial distances $\Delta x$ and $\Delta y$ and low differences in Doppler values $\Delta v_r$. The whole neighborhood criterion can be expressed as:
\begin{align} \label{eq:step1_3}
\sqrt{\Delta x^2 + \Delta y^2 + \epsilon^{-2}_{v_r}\cdot\Delta v_r^2} < \epsilon_{xyv_r} \:\wedge\: \Delta t < \epsilon_t.
\end{align}
In this case $\epsilon_{v_r}$ and $\epsilon_{xyv_r}$ have a scaling effect rather than representing absolute maximum velocity or spatial distance thresholds as in conventional DBSCAN processing.
The scaling of ${v_r}$ allows for better tuning capabilities than, e.g., normalizing all values to the same range.
As amplitudes often have very high variations even on a single object, they are completely neglected during clustering.
The time $t$ is not included in the Euclidean distance, i.e., $\Delta t$ is always required to be smaller or equal than its corresponding threshold $\epsilon_t$ due to real-time processing constraints.
In offline processing, this is addressed by using a sliding window of \SI{250}{\milli\second} length and an update rate of \SI{50}{\milli\second}.
All tuning parameters are adjusted using Bayesian Optimization \cite{Mockus1974} and a $V_1$ measure \cite{rosenberg2007} optimization score.
More details on $V_1$ are given in Sec.~\ref{sec:res_clustering}.

\subsection{Feature Selection And Extraction} \label{sec:meth_feat_sel}
For feature extraction, all labeled cluster sequences are first sampled in time using a non-overlapping sliding window of \SI{150}{\milli\second}.
The feature extraction window is chosen smaller than during the clustering process because this allows the subsequent classifier to better capture the variation between subsequent time frames.
In the next step, the features are extracted from each of the cluster samples so that with this increased number of feature vectors, the classifier can learn from more data.
The extracted features can be roughly divided into six groups.
The first four groups contain statistical values such as the minimum and maximum, the spread, and the standard deviation of the four base units (range, angle, amplitude, and Doppler).
The fifth group consists of geometric features describing the spatial distribution of detections in a cluster sample, e.g., the circularity or the size of a convex hull.
The final group addresses the ``micro-Doppler'' characteristics, i.e., the distribution of Doppler values within a cluster.
In total, 98 features are extracted from each cluster.
The entire list can be found in \cite{Scheiner2019IV}.
From the total of 98 features, only a subset is passed to each of the models in the following classifier ensemble.
Every classifier in that stage has its own task, hence, a different feature subset results in optimal performance.
Finding the exact optimal feature set for each classifier is an NP-hard problem.
Therefore, a guided backward elimination algorithm is used.
Backward elimination is a \emph{wrapper} method which repeatedly tests the utilized classifier and then eliminates the least fitting feature in a greedy fashion until a stopping criterion is reached \cite{Kohavi1997}.
A complete backward elimination run for all 98 features is computationally too expensive for each classifier in the ensemble.
Thus, for the backward elimination every feature is only assessed once but in a fixed order.
After each examination a feature is either dropped or kept which drastically reduces the computational effort.
The evaluation order is determined by the combination of two other feature selection techniques: the Joint Mutual Information (JMI) \cite{yang1999} and the Relief-based MultiSURF algorithm \cite{URBANOWICZ2018189}.
Both algorithms belong to the group of \emph{filters}.
As filter methods do not require multiple classifier trainings, they are usually much faster than wrapper methods, even though the latter often show superior results.
The combined approach of filter and wrapper methods is a compromise which yields well performing feature sets at a reasonable computational effort.

\subsection{Classification}
The classifier units used in this article are long short-term memory (LSTM) cells.
LSTMs are a special kind of recurrent neural network which introduce gating functions in order to avoid the vanishing gradient problem during training \cite{hochreiter97}.

A fixed configuration of 80 LSTM cells followed by a softmax layer is used for all classifiers in the ensemble.
The LSTM network is configured to accept up to eight consecutive feature vectors from the same cluster instance, if available.

The performance of the LSTM network is further improved by adding a few tweaks to the standard implementation: 
\emph{Multiclass binarization} is a wide-spread technique for improving the classification performance on moving road users, especially for unbalanced data sets.
The classification stage uses a combined one-vs-one (OVO) and one-vs-all (OVA) approach.
Class membership is estimated by summing all pairwise class posterior probabilities $p_{ij}$ from corresponding OVO classifiers.
Additionally, each OVO classifier is weighted by the sum of corresponding OVA classifier outputs $q_{i}$.
Subscripts $i$ and $j$ denote the corresponding class identifiers for which the classifier is	 trained.
During testing, this limits the influence of OVO classifiers which were not trained on the same class as the regarded sample, i.e., the OVA classifiers act as \emph{correction classifiers} \cite{Moreira1998}.
The final class decision for a feature vector $\bm{x}$ is then calculated as:
\begin{align} \label{eq:opc}
\texttt{id}(\bm{x}) = \argmax_{i \in \{1,...,K\}} \text{\hspace{1mm}} \sum_{j=1, j\neq i}^K p_{ij}(\bm{x}) \cdot (q_{i}(\bm{x}) + q_{j}(\bm{x})),
\end{align}
where $K$ is the number of classes in the training set.
This combined approach of OVO and OVA yields a total of $K(K+1)/2$ classifiers and feature sets, also indicated in Fig.~\ref{fig:pipeline}.
Moreover, during training, \emph{class weighting} is used to further reduce data imbalance effects.
Therefore, the influence of all training samples is adjusted inversely to their share in the total class distribution.
More details on the classification network and the multiclass binarization techniques for moving road users are given in \cite{Scheiner2018}.

\section{Results}
For the evaluation of the sensors, based on both data sets, each stage of the object detection framework is at first evaluated individually.
This is done in a ceiling analysis fashion, i.e., all steps are first computed assuming perfect accuracy of all other components.
Then, a combined result is estimated by evaluating the framework as a whole.
A summary of all results is presented in Tab.~\ref{tab:result_summary}.
All reported results are calculated based on distinct test sets which are not used for hyperparameter tuning or model training.
The two test sets consist of roughly \SI{20}{\percent} of their corresponding data set.
The split is sequence-based in order to avoid having the same object instances in the training and test split.
To this end, several million random permutations of all recorded sequences were tested for sequence combinations that yield class proportions similar to the full data set but at only \SI{20}{\percent} of their size.
At the feature extraction stage, the results are the found parameter sets and their degree of equivalence. 
Hence, the training set, i.e., the remaining \SI{80}{\percent} of the data is used for the evaluation at this stage.

\subsection{Clustering} \label{sec:res_clustering}
An important step towards a meaningful cluster evaluation is the choice of a good optimization score.
Compared to other clustering applications, this method is used to identify object instances and separate them from background points.
It is important to represent every road user with an individual cluster containing as many points as possible from the original one.
Additionally, the clustering process needs to stop before merging clusters from different road users or before adding background points to the object instance.
The majority of the data points in a radar scene are background detections.
Clusters containing only background detections and clutter are not desired in this application, but also not critical because the classifier stage should be able to distinguish road users from such unwanted cluster formations.

The V-measure \cite{rosenberg2007} combines two intuitive clustering criteria, \emph{homogeneity} and \emph{completeness}.
Completeness aims to assign all points from a single ground truth cluster into a single cluster prediction.
Contrary to that, homogeneity is maximal when a predicted cluster only contains points from a single ground truth cluster. 
$V_1$ is the harmonic mean of homogeneity and completeness:
\begin{equation}\label{eq:vmeas}
V_1 = 2\cdot \frac{\text{Homogeneity}\cdot\text{Completeness}}{\text{Homogeneity}+\text{Completeness}}.
\end{equation}
To stop the penalization of background clusters creation, the completeness score is calculated assuming perfect matching of the detections that belong to a labeled object in the ground truth.
This adaptation makes the score's objective sufficiently similar to the requirements for automotive radar clustering.

Four configurations are evaluated:
Both data sets are first optimized and evaluated individually ($A_{\text{\dsA},\text{\dsA}}$ and $A_{\text{\dsB},\text{\dsB}}$).
Then, the optimal configurations for both data sets are used to evaluate the other one ($A_{\text{\dsA},\text{\dsB}}$ and $A_{\text{\dsB},\text{\dsA}}$).
This gives a deeper insight into how well the data sets allows for generalization.
The results are listed in Tab.~\ref{tab:result_summary}.
$A_{\text{\dsB},\text{\dsB}}$ has the highest $V_1$ score of \SI{86.38}{\percent} and $A_{\text{\dsA},\text{\dsA}}$ scores much lower with $V_1=\SI{69.69}{\percent}$.
Similar to the scores, the cluster parameters differ a lot.
The best configuration for $\mathbb{S}_1$ has a setting of $N_{\text{min},\SI{50}{\meter}}=3$,
$\alpha_r=0.91$,
$\epsilon_{xyv_r}=\SI{1.4}{\meter}$,
$\epsilon_{v_r}=\SI{8.2}{\per\meter\per\second}$, and
$v_{r,\text{min}}=\SI{0.11}{\meter\per\second}$,
whereas $\mathbb{S}_2$ uses the following parameters:
$N_{\text{min},\SI{50}{\meter}}=3$,
$\alpha_r=0.5$,
$\epsilon_{xyv_r}=\SI{0.4}{\meter}$,
$\epsilon_{v_r}=\SI{2.4}{\per\meter\per\second}$, and
$v_{r,\text{min}}=\SI{0.63}{\meter\per\second}$.
It is hence possible to use much smaller $\epsilon$ regions for data of $\mathbb{S}_2$ which is apparently  beneficial for the overall performance.
Unfortunately, the largely different cluster parameterizations lead to massive decreases in the scores (both around \SI{20}{\percent}) when using them to segment the respective other data set, i.e., generalization is very low.

\subsection{Feature Selection}
\begin{figure}[tb]
	\centering
	\includegraphics{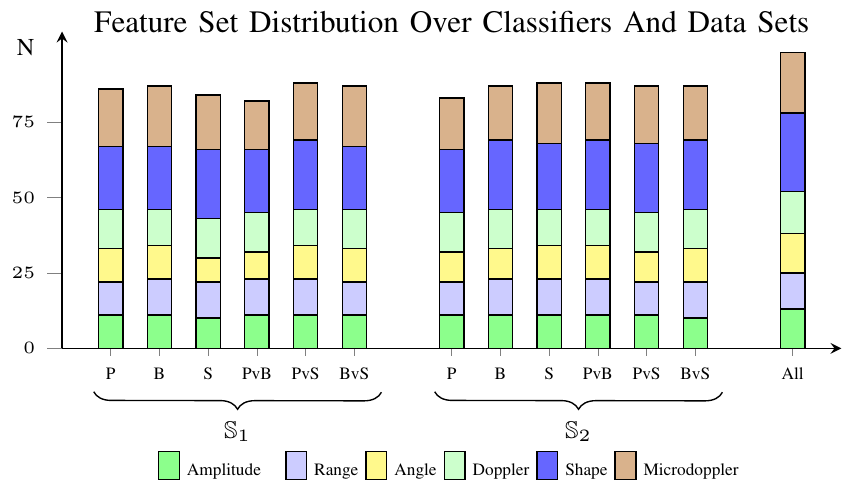}
	\caption{Feature distribution over classifiers in the ensemble. OVA classifiers are identified by their corresponding ``one'' class, i.e., pedestrian (P), bicycle (B), or static/garbage (S). OVO classifiers are indicated likewise by two letters. The full feature set distribution is displayed on the right for comparison.}
	\label{fig:feat_dist}
\end{figure}

When comparing the feature selection stage for both data sets, three factors are of interest for this evaluation:
First, does any of the two data sets require substantially more or less features?
Second, are specific features more important for one data set as for the other?
Third, how similar are the optimized feature sets of one sensor to the ones of the other sensor?
Fig.~\ref{fig:feat_dist} aims to answer the first two questions.
The diagram shows the amount of features presented to each classifier in the ensemble separately for both data sets.
Furthermore, the features are grouped into the six categories mentioned in Sec.~\ref{sec:meth_feat_sel}.
In comparison to previous studies (e.g. \cite{Scheiner2019IV}), the total amount of utilized features $N$ is rather high (averages: $|N_{\mathbb{S}_1}|=85.7$ and $|N_{\mathbb{S}_2}|=86.7$ out of $98$ in total).
However, this number remains more or less constant over different classifiers and data sets.
Also, there is no category for which a clear preference is visible.
The degree of equivalence can be estimated using the Jaccard index (aka. intersection over union -- IoU) between matching classifiers for both data sets.
The amount of common features for each classifier pair is divided by their union yielding the results in Tab.~\ref{tab:feat_sel}.
This summarizes to a mean IoU of \SI{89.7}{\percent} with \SI{2.5}{\percent} standard deviation which shows great conformity of both feature sets.
Thus, it is concluded that the choice of data set corresponding to a specific sensor has no remarkable impact on the feature extraction stage.
\begin{table}[tb]
	\renewcommand{\arraystretch}{1.3}
	\caption{Jaccard index between each pair of classifiers in the ensembles for both data sets.  OVA classifiers are identified by their corresponding ``one'' class, i.e., pedestrian (P), bicycle (B), or static/garbage (S). OVO classifiers are indicated likewise by two letters.}
	\label{tab:feat_sel}
	\centering
	\resizebox{0.98\columnwidth}{!}{%
	\begin{tabular}{ccccccc}
		\toprule
		P & B & S & PvB & PvS & BvS & Mean \& StdDev \\ \midrule
		\SI{85.7}{\percent} & \SI{91.2}{\percent} & \SI{91.1}{\percent} & \SI{86.8}{\percent} & \SI{92.3}{\percent} & \SI{91.2}{\percent} & \SI{89.7}{\percent} $\pm$ \SI{2.5}{\percent}\\
		\bottomrule
	\end{tabular}%
	}
\end{table}

\subsection{Classification}
\begin{figure}[tb] 
	\centering
	\includegraphics{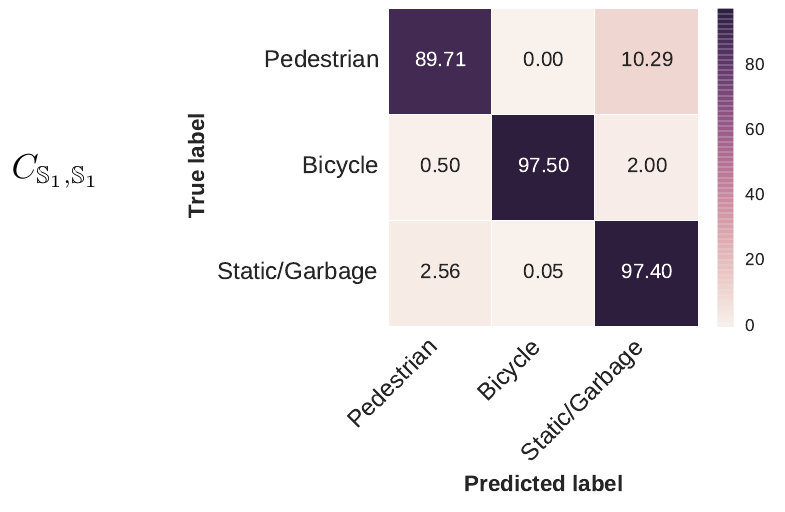}

	\vspace{10pt}
	\includegraphics{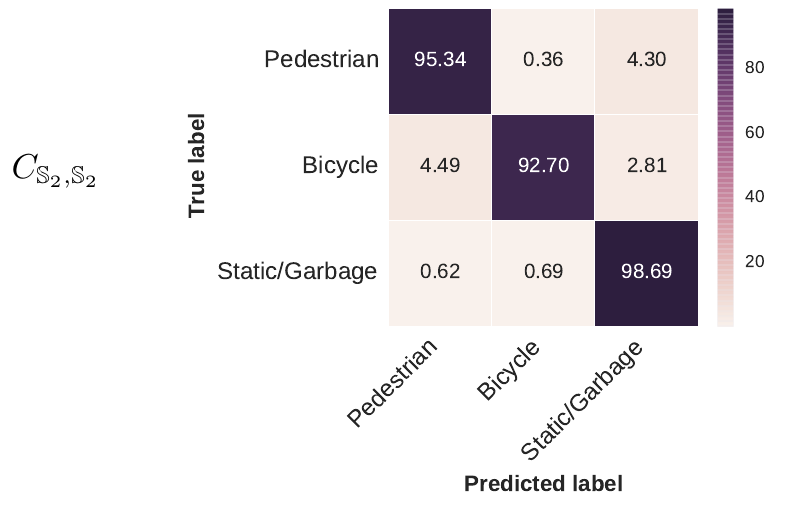}
	\caption{Confusion matrices of the two best performing classification experiments. On top: $C_{\text{\dsA},\text{\dsA}}$, and on the bottom: $C_{\text{\dsB},\text{\dsB}}$.}
	\label{fig:cm} 
\end{figure}

As mentioned before, both data sets contain strong imbalances between classes.
To preserve the influence of each individual class, all classification scores are reported as \emph{macro-averaged $F_1$ scores}.
The $F_1$ score is the harmonic mean of precision (true positive / predicted positive) and recall (true positive / condition positive).
Macro-averaging uses the mean value of all $K$ classes' individual $F_1$ scores:
\begin{equation}  \label{eq:macroavg}
F_{1,\text{macro}}=\frac{1}{K}\sum_{i=1}^K F_{1,i}.
\end{equation}
All classification results are listed in Tab.~\ref{tab:result_summary}.
The model trained and tested on $\mathbb{S}_2$ performs best, with a score of $C_{\text{\dsB},\text{\dsB}}=\SI{95.80}{\percent}$ compared to $C_{\text{\dsA},\text{\dsA}}=\SI{95.46}{\percent}$.
The \SI{0.34}{\percent} gain in $F_1$ at total score of \SI{>95}{\percent} would be a good result if the data sets contained identical scenarios.
As this is not the case, the difference has to be regarded too low to make the strong claim that the classifier generally benefits from the increased sensor resolution.
In order to look closer into the classification results, Fig.~\ref{fig:cm} shows the confusion matrices for the two single data set experiments.
When comparing both matrices, it is apparent that even though the $F_1$ scores for both experiments are not remarkably different, the formation of those scores is.
The most decisive factor is that for $C_{\text{\dsA},\text{\dsA}}$, the confusion of both vulnerable road user (VRU) classes with the background class is much higher than for $C_{\text{\dsB},\text{\dsB}}$.
In turn, the latter has a higher confusion between VRUs.
In practice, the second behavior is more desirable as VRUs are not overlooked.
In terms of generalization between data sets, the classification stage behaves similar to the clustering stage.
While $C_{\text{\dsB},\text{\dsA}}$ still manages to make some useful predictions, $C_{\text{\dsA},\text{\dsB}}$ cannot overcome the class imbalance and simply predicts the \emph{static/garbage} for almost all samples resulting in $F_1=\SI{28.34}{\percent}$.
This leads to the conclusion, that the features from the previous stage, even though important for both data sets, are not robust enough to cover the difference in data.
This may be the case, e.g., for radar cross section (amplitude) estimates which are often part of the internal sensor processing, i.e., the processing may differ between sensor types.
Another example is the number of detection points in a cluster which varies heavily between sensors.
The coverage of all specific features is, however, not in the scope of this article.
Summarizing the classification stage, it can be stated that there seem to be some small benefits in classification performance when using a high resolution radar.
However, the improvements are less distinct as for the clustering stage.

\subsection{Object Detection Framework}
To evaluate the whole object detection pipeline at once, new metrics are necessary.
The important difference to previous steps is that only fractions of objects may be classified correctly or different object instances might be merged during clustering.
As the choice of evaluation metric has a big influence on the final results, four different metrics are proposed.
\paragraph{Point-wise $F_1$ score}
Adopted from the classification stage, a macro-averaged $F_1$ score is calculated, this time based on the prediction of all detection points instead of cluster samples.
The advantage of this score is that it gives comprehensible feedback about how well the scene was segmented.
\paragraph{Instance-based $F_1$ score}
While the simplicity of the first score may be advantageous to gain a general understanding, it does not capture correct instance segmentation.
In image-based object detection, the usual way to decide if an object is detected or not is by calculating the pixel-based IoU \cite{everingham_pascal_2015}.
This can easily be adopted to radar point clouds by calculating the intersection and union based on detection points instead of pixels:
\begin{equation}
\textit{IoU} = \frac{\abs{\text{predicted detections} \cap \text{true detections}}}{\abs{\text{predicted detections} \cup \text{true detections}}} \,\text{.}
\end{equation}
A VRU instance is defined as correctly detected if the cluster's IoU is greater or equal to $0.5$ for a ground truth instance with the same label.
This corresponds to a true positive ($\textit{TP}$).
Other predicted instances on the same ground truth object make up false positives ($\textit{FP}$).
Finally, non-detected VRU ground truth instances count as false negatives ($\textit{FN}$) and everything else as true negatives ($\textit{TN}$).
By using an alternative notation of $F_1$, 
\begin{equation}
F_1 = \frac{2 \textit{TP}}{2 \textit{TP} + \textit{FP} + \textit{FN}},
\end{equation}
this can easily be used to calculate an instance-based $F_1$ score, which is also macro-averaged according to Eq.~\ref{eq:macroavg}.
Most recently, this metric was also used and illustrated in \cite{Palffy2020}.
\paragraph{Binary VRU detection score}
The second criterion can be eased in order to not punish the object detector if it correctly segments an object, but assigns the wrong VRU label, i.e., \emph{pedestrian} instead of \emph{bicycle} or the other way round.
In this case all cluster samples with $\textit{IoU}\geq0.5$ for any VRU object count as \textit{TP} if either VRU label is predicted.
If one is only interested in how well road users are recognized by the proposed object detector, a VRU-based true positive rate (TPR or recall), can be calculated as:
\begin{equation}
\textit{TPR} = \frac{\textit{TP}}{\textit{TP}+\textit{FN}}.
\end{equation}
On its own, the recall can be easily misleading, since it does not account for $\textit{FP}$.
In addition to the other scores, however, this gives extra information about where fine-tuning of the object detector is most appropriate.
\paragraph{VRU Balanced Accuracy}
A more general version of a VRU-based score is the Balanced Accuracy (BAAC), which is calculated as:
\begin{equation}
\textit{BAAC} = \frac{\textit{TPR}+\textit{TNR}}{2},
\end{equation}
where the true negative rate $\textit{TNR}=\nicefrac{\textit{TN}}{\textit{TN}+\textit{FP}}$ also takes into account the performance of the background class rejection.
Similar to the $F_1$ score, BAAC is an indicator for classification or object detection tasks, which works particularly well on imbalanced data sets.

All results are presented in Tab.~\ref{tab:result_summary}.
It is clearly visible, that the next generation radar in experiment $M_{\text{\dsB},\text{\dsB}}$ outperforms the conventional sensor in $M_{\text{\dsA},\text{\dsA}}$ with big margins in all categories.
Despite their similarity in terms of pure classification performance on perfect data samples, the differences are very distinct on an object detection level.
Both $F_1$ scores improve roughly \SI{7}{\percent} and \SI{10}{\percent}, the $\textit{TPR}$ even goes up by \SI{20}{\percent}.
This is a strong indicator that the improved results at the clustering stage are extremely beneficial for the overall object detection performance.
Only, the generalization ability for the cross data set experiments $M_{\text{\dsA},\text{\dsB}}$ and $M_{\text{\dsB},\text{\dsA}}$ is now close to non-existing.
When evaluating the two VRU-based scores in any data set combination, it is clearly visible that the \textit{TNR} part of BAAC is rather high compared to the \textit{TPR} part.
This stresses the advantage of having the recall as a separate score, as this is the one to optimize further.

\begin{table}[tb]
	\renewcommand{\arraystretch}{1.3}
	\caption{Evaluation result summary for all evaluated categories and experiments. Details on all scores are given in the corresponding sections.}
	\label{tab:result_summary}
	\centering
	\begin{tabularx}{\linewidth}{|Y|Y|Y|Y|}
		\hline 
		\multicolumn{2}{|c|}{\textbf{Clustering Results}} &  \multicolumn{2}{c|}{\textbf{Classification Results}}  \\ 
		\hline 
		$\mathbf{A}_\textbf{opt,eval}$ & \bfvscore & $\mathbf{C}_\textbf{train,eval}$ & \bffscore \\ 
		\hline 
		$A_{\text{\dsA},\text{\dsA}}$ & \SI{69.69}{\percent} & $C_{\text{\dsA},\text{\dsA}}$ & \SI{95.46}{\percent} \\ 
		\hline 
		$A_{\text{\dsB},\text{\dsB}}$ & \SI{86.38}{\percent} & $C_{\text{\dsB},\text{\dsB}}$ & \SI{95.80}{\percent} \\ 
		\hline 
		$A_{\text{\dsB},\text{\dsA}}$ & \SI{51.56}{\percent} & $C_{\text{\dsB},\text{\dsA}}$ & \SI{50.71}{\percent} \\ 
		\hline 
		$A_{\text{\dsA},\text{\dsB}}$ & \SI{63.89}{\percent} & $C_{\text{\dsA},\text{\dsB}}$ & \SI{28.34}{\percent} \\ 
		\hline 
	\end{tabularx}   
	\begin{tabularx}{\linewidth}{|c|Y|c|c|Y|}	
		\multicolumn{5}{|c|}{\textbf{Combined Detection Results}} \\ 
		\hline	
		$\mathbf{M}_\textbf{train,eval}$ & \textbf{Point} \bffscore & \textbf{Instance} \bffscore & $\textbf{TPR}_\textbf{VRU}$ &  $\textbf{BAAC}_\textbf{VRU}$ \\ 
		\hline 
		$M_{\text{\dsA},\text{\dsA}}$ & \SI{77.61}{\percent} & \SI{76.08}{\percent} & \SI{56.49}{\percent} & \SI{76.61}{\percent} \\ 
		\hline 
		$M_{\text{\dsB},\text{\dsB}}$ & \SI{88.03}{\percent} & \SI{83.39}{\percent} & \SI{77.17}{\percent} & \SI{82.83}{\percent} \\ 
		\hline 
		$M_{\text{\dsA},\text{\dsB}}$ & \SI{44.94}{\percent} & \SI{24.62}{\percent} & \SI{19.18}{\percent} & \SI{50.00}{\percent} \\  
		\hline 
		$M_{\text{\dsB},\text{\dsA}}$ & \SI{8.25}{\percent} & \SI{0.0}{\percent} & \SI{0.0}{\percent} & \SI{50.07}{\percent}
		\\
		\hline
	\end{tabularx} 
\end{table}

\section{Discussion And Conclusion}
In this article, two data sets from different radar sensor generations have been tested against each other.
For this purpose, properties for a fair comparison of two radar data sets have been worked out and applied to crop two proprietary data sets to comparable subsets.
Both subsets are processed using an identical radar object detection framework consisting of a clustering algorithm, a feature selection stage, and a recurrent neural network ensemble.
The interaction of all modules in the framework is an important part of the evaluation, as to date, most research has been focused solely on single components.
Results are reported for each intermediate category as well as for the whole framework.
The main question: \emph{``Does an object detector benefit from next generation radar sensors?''}, can be positively answered.
The instance-based $F_1$ object detection score improves from \SI{76.1}{\percent} with an off-the-shelf sensor to \SI{83.4}{\percent} using a high resolution next generation radar.
Even though, these numbers do not seem very high when compared to modern image-based object detectors, it is an excellent result for solely radar-based VRU detection.
The greatest improvements are made in the clustering stage.
Formerly the weak point of the pipeline, the clustering errors are now at a mediocre level.
The main reason for this improvement is that for the next generation radar, the detection points on an object are located close enough for the clustering algorithm to utilize much smaller neighborhood regions. 
This effectively allows for better separation of real objects and background clutter.
In the classification stage, only a slight improvement in $F_1$ scores is obtained. Regarding the confusion between classes, VRU discrimination is a lot better with the new sensor technology.
While no big differences can be observed in the feature extraction module, tests at all other stages of the framework showed that the generalization capability from one sensor to the other is minimal.
Nevertheless, it is highly likely -- and shall be tested in future work -- that transfer learning, i.e., using either one data set for pre-training a classifier and fine-tuning it with the other data set, leads to an improved performance.
Furthermore, the weak generalization capability without fine-tuning motivates the search for more robust features.
As an example, a principal component analysis could be used to obtain lower-dimensional features with a possibly higher generalization capability.
Another approach is adjusting existing features by calibrating them with sensor properties, e.g., the number of detection points in a cluster may be scaled by the maximum possible number of points during one sensor scan.
This might be of even greater interest in cases where different types of sensors, e.g., short-range radars and long-range radars, complement each other in the same vehicle.


\bibliographystyle{IEEEtran}
\bibliography{IEEEabrv,mybibfile}

\end{document}